\documentclass[conference]{IEEEtran}
\IEEEoverridecommandlockouts
\usepackage{cite}
\usepackage{amsmath,amssymb,amsfonts}
\usepackage{algorithmic}
\usepackage{graphicx}
\usepackage{textcomp}
\usepackage{xcolor}
\usepackage{xspace}
\usepackage{subfigure}
\usepackage{esvect}
\usepackage{makecell,rotating}
\usepackage{booktabs}
\usepackage{multirow}
\usepackage{url} 
\usepackage{colortbl} 
\def\BibTeX{{\rm B\kern-.05em{\sc i\kern-.025em b}\kern-.08em
    T\kern-.1667em\lower.7ex\hbox{E}\kern-.125emX}}

\newcommand{\ie}{\textit{i.e.}\xspace}
\newcommand{\eg}{\textit{e.g.}\xspace}
\newcommand{\etal}{\textit{et al.}\xspace}

\begin{document}

\title{ Shape Transformation Driven by Active Contour for Class-Imbalanced Semi-Supervised Medical Image Segmentation
}

\author{
\IEEEauthorblockN{1\textsuperscript{st} Yuliang Gu}
\IEEEauthorblockA{\textit{School of Computer Science} \\
\textit{Wuhan University}\\
Wuhan, China \\
yuliang\_gu@whu.edu.cn
}
\and

\IEEEauthorblockN{2\textsuperscript{nd} Yepeng Liu}
\IEEEauthorblockA{\textit{School of Computer Science} \\
\textit{Wuhan University}\\
Wuhan, China \\
yepeng.liu@whu.edu.cn}
\and
\IEEEauthorblockN{3\textsuperscript{rd} Zhichao Sun}
\IEEEauthorblockA{\textit{School of Computer Science} \\
\textit{Wuhan University}\\
Wuhan, China \\
zhichaosun@whu.edu.cn}
\and
\IEEEauthorblockN{4\textsuperscript{th} Jinchi Zhu}
\IEEEauthorblockA{\textit{School of Computer Science} \\
\textit{Wuhan University}\\
Wuhan, China \\
jinchi.zhu@whu.edu.cn}
\and
\IEEEauthorblockN{5\textsuperscript{th} Yongchao Xu* } 
\IEEEauthorblockA{\textit{School of Computer Science} \\
\textit{Wuhan University}\\
Wuhan, China \\
yongchao.xu@whu.edu.cn}
\and
\IEEEauthorblockN{6\textsuperscript{th} Laurent Najman}
\IEEEauthorblockA{
\textit{Univ Gustave Eiffel}\\
\textit{CNRS, LIGM} \\
Marne-la-Vall$\acute{e}$e, France \\
laurent.najman@esiee.fr}

\thanks{* Corresponding Author}
}

\maketitle

\begin{abstract}
 Annotating 3D medical images demands expert knowledge and is time-consuming. As a result, semi-supervised learning (SSL) approaches have gained significant interest in 3D medical image segmentation. The significant size differences among various organs in the human body lead to imbalanced class distribution, which is a major challenge in the real-world application of these SSL approaches. To address this issue, we develop a novel \textbf{S}hape \textbf{T}ransformation driven by \textbf{A}ctive \textbf{C}ontour (STAC), that enlarges smaller organs to alleviate imbalanced class distribution across different organs. Inspired by curve evolution theory in  active contour methods, STAC employs a signed distance function (SDF) as the level set function, to implicitly represent the shape of organs, and deforms voxels in the direction of the steepest descent of SDF (\ie, the normal vector). To ensure that the voxels far from expansion organs  remain unchanged, we design an SDF-based weight function to control the degree of deformation for each voxel. We then use STAC as a data-augmentation process during the training stage. Experimental results on two benchmark datasets demonstrate that the proposed method significantly outperforms some state-of-the-art methods. Source code is publicly available at https://github.com/GuGuLL123/STAC.
\end{abstract}


\section{Introduction}
Precise segmentation of medical images is crucial for computer-aided diagnosis (CAD) systems. While supervised segmentation approaches have demonstrated remarkable success with extensive labeled datasets, the process of manual segmentation remains laborious and time-consuming. Recently, semi-supervised segmentation techniques have attracted considerable interest for utilizing easily accessible unlabeled images to enhance the precision of segmentation models. 
These methods commonly leverage two strategies: consistency regularization and pseudo labeling. 
Consistency regularization approaches are mainly based on some form of smoothness assumption~\cite{chen2022semi}, that aims to produce consistent  results under small perturbations at data level~\cite{sajjadi2016regularization,you2022simcvd,suzuki2020adversarial,miyato2018virtual} and model level~\cite{rasmus2015semi,park2018adversarial,tarvainen2017mean,wu2022mutual,qiao2018deepcotraining}.  
Pseudo labeling approaches~\cite{chen2021semi,lyu2022pseudo,qiao2022semi} leverage predictions of the model on unlabeled data to generate pseudo labels, thereby expanding the initial labeled dataset.

\begin{figure*}[t]
\centering
\subfigure[Image]{
\begin{minipage}[b]{0.16\linewidth}    %
\includegraphics[width=1\linewidth]{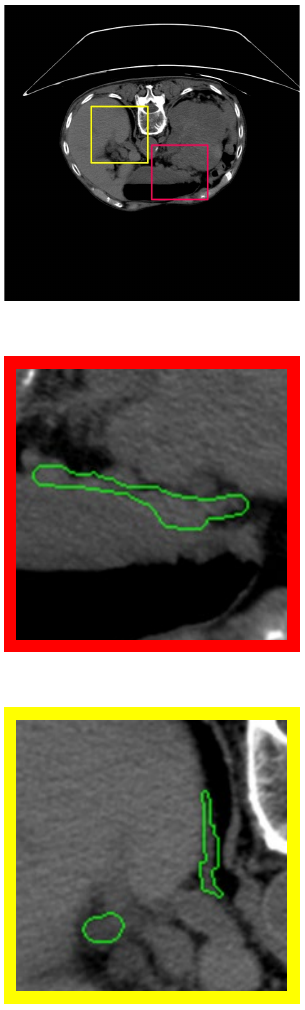}
\end{minipage}}\hspace{1.5mm}
\subfigure[Label]{
\begin{minipage}[b]{0.16\linewidth}
\includegraphics[width=1\linewidth]{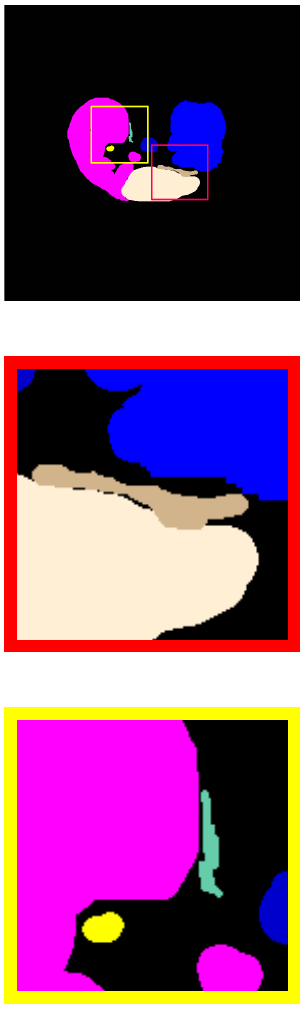}
\end{minipage}}\hspace{1.5mm}
\subfigure[Transformed Image]{
\begin{minipage}[b]{0.16\linewidth}
\includegraphics[width=1\linewidth]{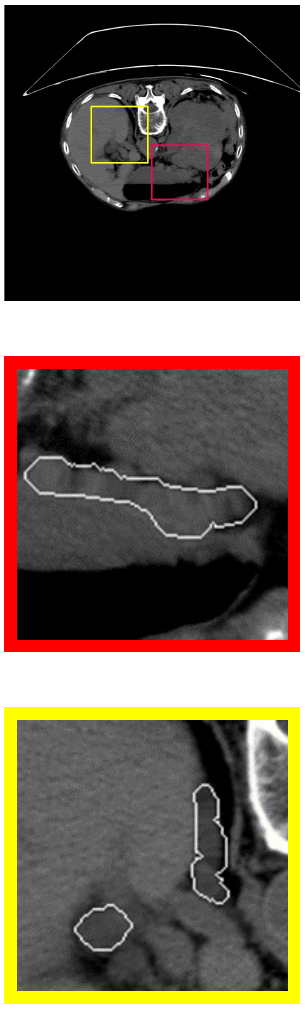}
\end{minipage}}\hspace{1.5mm}
\subfigure[Transformed Label]{
\begin{minipage}[b]{0.16\linewidth}
\includegraphics[width=1\linewidth]{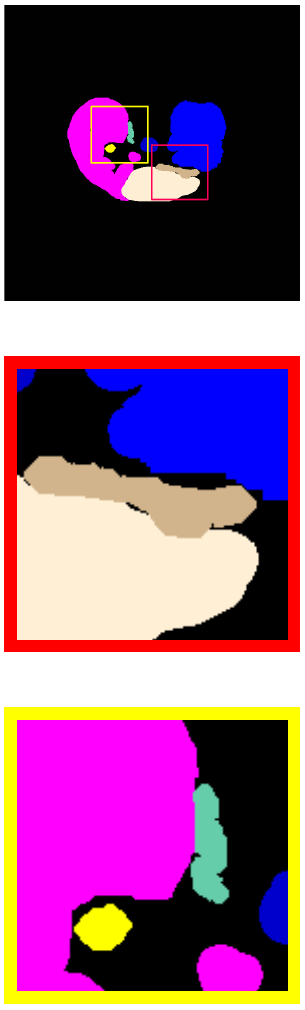}
\end{minipage}}\hspace{1.5mm}

\caption{Motivation of STAC:  enlarging smaller organs helps to alleviate imbalanced class distribution across different organs. The second and third rows are magnified images of the corresponding color boxes in the first row. The green curve in the Image represents the corresponding edges of the ground truth for categories with smaller pixel proportions. The white curve in the Transformed Image represents the edges of the corresponding ground-truth category, after processing with STAC.}
\label{fig:motivation}
\end{figure*}

The size of human organs can vary considerably, frequently causing a notable imbalance in the number of voxels among different categories (as shown in Fig.~\ref{fig:motivation}). Recently, some works address this issue of imbalanced class distribution in semi-supervised medical image segmentation by improving the network's representation ability of imbalanced classes and refining their pseudo labels. 
For instance, Basak \etal~\cite{basak2022addressing} propose a dynamic learning strategy that tracks class-wise confidence during training and incorporates fuzzy fusion and robust class-wise sampling to improve segmentation performance for under-represented classes. To address the issue that the proposed model in~\cite{basak2022addressing} can not well model the difficulty, DHC~\cite{wang2023dhc} introduces Distribution-aware Debiased Weighting and Difficulty-aware Debiased Weighting, two strategies that use pseudo labels to dynamically address data and learning biases. 
Yuan \etal~\cite{yuan2023semi}  combine a traditional linear-based classifier with a prototype-based classifier to  alleviate the class-wise bias exhibited in each individual sample. 
However, these methods do not address the imbalanced class issue from an intuitive perspective in data level, that is, some organs are too small, resulting in low number of voxels.

In this paper, we enlarge smaller organs to alleviate the issue of imbalanced class distribution across different organs. In active contour methods, the contours evolve in the direction of the normal vector and are represented implicitly by a level set function to ensure stability during the evolution process. Inspired by this, we propose the Shape Transformation driven by Active Contour (STAC) method. STAC method utilizes a signed distance function (SDF) as the level set function to subtly represent the shapes of organs. It adjusts the voxels towards the direction of the steepest descent in SDF, which corresponds to the normal vector of the shape in zero level set. To ensure that voxels distant from the expanding organs remain unaffected, we develop a weight function based on SDF to carefully regulate the extent of deformation for each voxel. As shown in Fig.~\ref{fig:motivation}, we generate pairs of images and annotations (pseudo-labels for unlabeled images) that alleviate the class imbalance issue through our STAC method. Owing to the fact that evolution along the direction of the normal vector is less likely to produce distorted shapes, STAC can generate stable shape transformations, making the generated image more realistic. Using STAC for data-augmentation during training enhances the network's ability to represent categories that have a smaller number of voxels. Experimental results demonstrate that enlarging smaller organs to alleviate the issue of imbalanced class distribution is effective. Our experiments demonstrate that the proposed STAC method significantly outperforms some state-of-the-art methods on two datasets.

\begin{figure*}[!t]
\centering
\includegraphics[width = 1\linewidth]{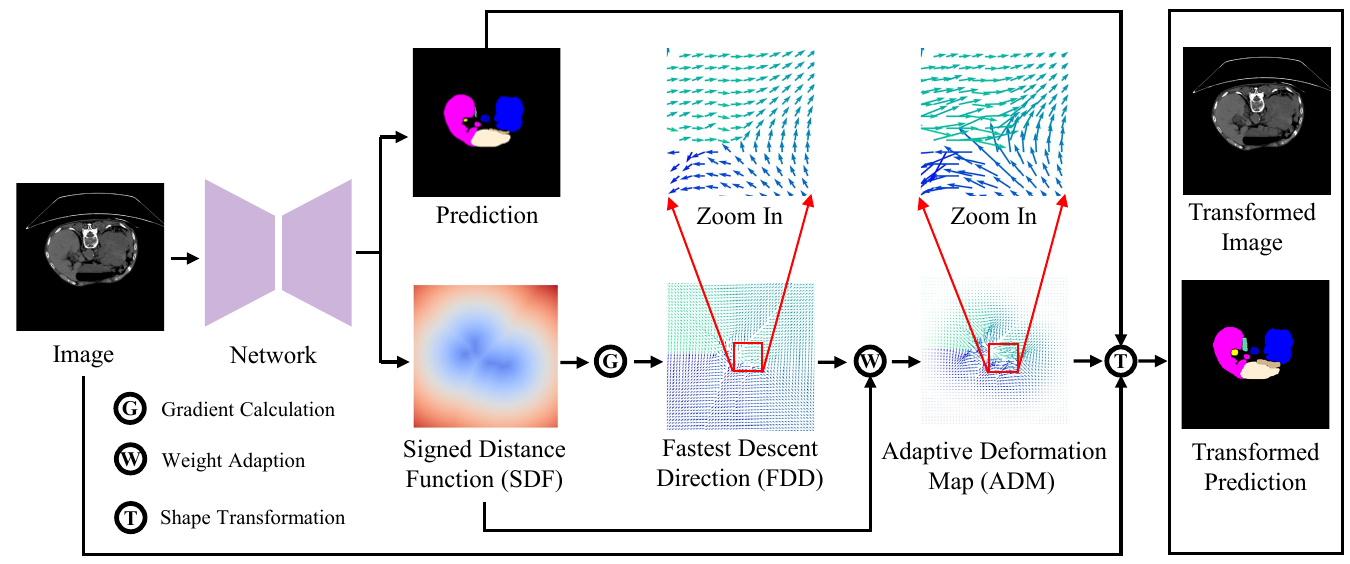}
\caption{The pipeline of the proposed STAC framework, that is used for data-augmentation. For unlabeled samples, the Fastest Descent Direction (FDD) is defined as the gradient of the Signed Distance Function (SDF), which represents the direction of Adaptive Deformation Map (ADM). The degree of the ADM is SDF-based weight function. 
The shape transformation for enlarging smaller organs is obtained by interpolating the unlabeled images and pseudo labels according to the ADM. 
For labeled samples, the ADM is directly obtained using the label.
} 
\label{fig:pipeline}
\end{figure*}
\section{Related Works}
\subsection{Semi-supervised medical image segmentation}

Semi-supervised learning (SSL) utilizes limited annotated data alongside abundant unlabeled data to enhance model performance. SSL methods are mainly categorized into self-training~\cite{grandvalet2005semi}, where pseudo labels are assigned to unlabeled data for retraining, and consistency regularization~\cite{laine2016temporal, tarvainen2017mean}, which relies on the assumption that model outputs should remain stable under some perturbations~\cite{chen2022semi}. Image-level perturbations include simple random augmentations~\cite{sajjadi2016regularization} and adversarial approaches~\cite{ miyato2018virtual}, while model-level perturbations involve direct stochastic modifications (\eg, Gaussian noise~\cite{rasmus2015semi} or dropout~\cite{park2018adversarial}), parameter ensembling via exponential moving averages~\cite{tarvainen2017mean}, or generating variations through different decoders or architectures~\cite{wu2022mutual, qiao2018deepcotraining}.

Semi-supervised learning is extensively employed in medical image segmentation, to ease the challenges of manual annotation. Consistency regularization methods~\cite{basak2023pseudo, lei2022semi, jin2022semi, xiang2022fussnet, basak2022addressing, wang2023mcf, lyu2022pseudo} have significantly enhanced semi-supervised medical semantic segmentation, utilizing strategies such as the Mean Teacher framework~\cite{tarvainen2017mean} or Co-training~\cite{qiao2018deepcotraining} to introduce model-level variations. Another way involves creating diverse versions of the same image to enforce prediction consistency across image variations~\cite{huang2022semi, xu2021shadow, fan2022ucc, peiris2021duo, wang2023cat}. A prevalent method for image variations is the weak-to-strong paradigm~\cite{fan2022ucc}, which utilizes images ranging from weakly to strongly augmented to promote model consistency. 
Several methods utilize adversarial training strategies to create adversarial perturbations on images, thereby enhancing the robustness of predictions against these perturbations~\cite{peiris2021duo, wang2023cat}. Additionally, an increasing number of techniques improve model performance by employing pseudo labels for training unlabeled images~\cite{lyu2022pseudo, qiao2022semi}. 
The presence of noisy labels in pseudo labels for unlabeled images necessitates a careful assessment of their confidence levels~\cite{qiao2022semi, wang2021semi}. Additionally, certain methodologies emphasize the rectification of pseudo labels during the training process~\cite{liu2022perturbed}. Beyond these strategies, other methods harness contrastive learning to attain consistent feature representation~\cite{you2022simcvd, basak2023pseudo}.

\subsection{Class-Imbalanced Learning}
Real-world datasets often exhibit a class-imbalanced label distribution, complicating the standard training and generalization of machine learning models~\cite{liu2019large}. Numerous algorithms~\cite{cao2019learning,shu2019meta,byrd2019effect,buda2018systematic} have been proposed to address this problem. The most prevalent approach for addressing class imbalances involves rebalancing the training objective based on class-specific sample sizes.  These methods are mainly categorized into re-weighting~\cite{cao2019learning,shu2019meta}, which impacts the loss function by assigning higher costs to examples from minority classes, and re-sampling~\cite{byrd2019effect, buda2018systematic}, which directly adjusts the label distribution by over-sampling the minority class or under-sampling the majority class, or both, to achieve a balanced sampling distribution.

The issue of class imbalance poses a significant challenge for extending existing SSL-based methods t   o more practical settings. 
CReST~\cite{wei2021crest} employs a self-training approach that enhances the SSL model by adaptively selecting pseudo-labeled data from the unlabeled set to augment the original labeled set. Unlike traditional self-training methods, CReST~\cite{wei2021crest} chooses pseudo-labels based on label frequency to progressively balance the class distribution, favoring predictions from minority classes. 
Lin \etal~\cite{lin2022cld} introduce the CLD method, which mitigates data bias by adjusting the overall loss function according to the voxel count of each class. Basak \etal~\cite{basak2022addressing} propose a dynamic learning strategy that monitors class-specific confidence throughout training, integrating fuzzy fusion and robust class-wise sampling to enhance segmentation performance for underrepresented classes. However, the model proposed in~\cite{basak2022addressing} does not consider the varying difficulty levels in modeling different categories. To address this issue, DHC~\cite{wang2023dhc} introduces two innovative strategies: Distribution-aware Debiased Weighting and Difficulty-aware Debiased Weighting. These approaches utilize pseudo labels to dynamically counteract data and learning biases.

\section{Method}

\subsection{Curve Evolution Theory of Active Contour Model}
A shape can be represented by its edge curves. Let $C : s\in [0,1] \rightarrow (x(s),y(s))\in \mathbb{R}^{2}$ denote a parametric curve. The outward normal of $C$ is $\vec{N}(s) = (y^{\prime}(s),-x^{\prime}(s))$. Many active contour methods~\cite{wang2019object,li2010distance} evolve curves according to a velocity vector in the direction of its normal, which is expressed as $ \frac{\partial C(x(s), y(s), t)}{\partial t} = V \vec{N}$. Level set method provides an effective and stable numerical solution for solving the above partial differential equation problem.
In the level set implementation, an active curve $C$ is implicitly represented as the zero level set of a function $\phi$, denoted as $C = \{(x,y)| \phi(x,y) = 0\}$. The evolution of the level set function is $\frac{\partial \phi(x, y, t)}{\partial t} = V || \nabla \phi||$.

Inspired by the evolution of curves along the normal vector and the level set representation, we design STAC to alleviate imbalanced class distribution. The pipeline of the proposed STAC is depicted in Fig.~\ref{fig:pipeline}.

\subsection{Shape Transformation Driven by Active Contour Evolution}
 Let $\mathcal{D} = \mathcal{D}_l\cup \mathcal{D}_{u}$ be the whole training dataset, where $\mathcal{D}_l$ and $\mathcal{D}_u$ represent labeled and unlabeled dataset.
For a labeled data pair $(X_{l},Y) \in \mathcal{D}_l$, we denote the minority classes as $M$. The organs we need to enlarge are $\mathcal{O} = \{(x,y,z)\in \mathbb{R}^{3} \;|\; Y(x,y,z) \in M \}$. We define the signed distance function as the level set representation of $\mathcal{O}$:
\begin{equation}\label{sdf}
\phi_{\mathcal{O}}(\textbf{p}) =\left\{\begin{aligned}
& -\inf _{\textbf{q} \in \partial \mathcal{O}}\|\textbf{p}-\textbf{q}\|_{2}, &\textbf{p} \in &\mathcal{O}_{\mathrm{in}} 
\\& 0, &\textbf{p} \in &\mathcal{\partial \mathcal{O}} 
\\& +\inf _{\textbf{q} \in \partial \mathcal{O}}\|\textbf{p}-\textbf{q}\|_{2},&\textbf{p} \in &\mathcal{O}_{\mathrm{out}}
\end{aligned}\right.
\end{equation}
where $\partial \mathcal{O}$ represents the surface of $\mathcal{O}$. $\mathcal{O}_{\mathrm{in}} $ and $\mathcal{O}_{\mathrm{out}} $ denote the inside region and outside region of $\mathcal{O}$.
The normal of $\phi_{\mathcal{O}}$ is denoted as $\vec{N}_{\mathcal{O}}(x,y,z) = (\frac{\partial \phi_{\mathcal{O}}}{\partial x}, \frac{\partial \phi_{\mathcal{O}}}{\partial y}, \frac{\partial \phi_{\mathcal{O}}}{\partial z})$. 
To ensure that voxels far from the expanding organs remain unaffected, we devise a weight coefficient $W_{\mathcal{O}}(x,y,z)$ based on SDF to carefully control the degree of deformation for each voxel.
\begin{equation}\label{eq:dis_weight}
W_{\mathcal{O}}(x,y,z) = \alpha \times e^{\beta \times \lvert \phi_{\mathcal{O}}(x,y,z)\rvert},
\end{equation}
where $\alpha$ and $\beta$ are hyper-parameters.
The adaptive deformation map $\vec{D}_{\mathcal{O}}(x,y,z)$ is obtained by multiplying the normal vector with a weight coefficient.
\begin{equation}\label{eq:final_dirction}
\vec{D}_{\mathcal{O}}(x,y,z) = W_{\mathcal{O}}(x,y,z) \times \vec{N}_{\mathcal{O}}(x,y,z).
\end{equation}

By transforming the shape of the organs along the direction of the normal vector of the level set function,  $X^{transformed}_{l}$ is denoted as:

\begin{equation}\label{shape_transform_image}
\begin{aligned}
X^{transformed}_{l}(x,y,z) = X_{l}(x & +\frac{\partial \phi_{\mathcal{O}}}{\partial x}\times W_{\mathcal{O}}, \\
y & +\frac{\partial \phi_{\mathcal{O}}}{\partial y}\times W_{\mathcal{O}}, \\ z & +\frac{\partial \phi_{\mathcal{O}}}{\partial z}\times W_{\mathcal{O}}). \\
\end{aligned}
\end{equation}

The same transformation is applied to generate $Y^{transformed}$. 
\begin{equation}\label{shape_transform_label}
\begin{aligned}
Y^{transformed}(x,y,z) = Y(x & + \frac{\partial \phi_{\mathcal{O}}}{\partial x}\times W_{\mathcal{O}}, \\
y & + \frac{\partial \phi_{\mathcal{O}}}{\partial y}\times W_{\mathcal{O}}, \\
z & +\frac{\partial \phi_{\mathcal{O}}}{\partial z}\times W_{\mathcal{O}}).
\end{aligned}
\end{equation}
For unlabeled image $X_{u} \in D_{u}$, pseudo SDF $\phi_{pre}$ is predicted by the network. We use $\phi_{pre}$  to calculate the $\vec{D}_{pre}(x,y,z)$ as adaptive deformation map. 
$X_{u}$ and its pseudo-label $\hat{Y}$ is transformed using $\vec{D}_{pre}(x,y,z)$.

\subsection{Network architecture and training process}
 Our method can be integrated into any existing semi-supervised method, in a plug-and-play manner. In this paper, we use mutual supervision of two networks trained with cross-entropy loss (Refer to Eq.\eqref{labeled superviseA} and \eqref{unlabeled superviseA} below for this classic approach.), as an illustration of our proposal. 
 For $i\in[1,2]$, let the feature extraction module be denoted as $f^i$ the classification head as $h^i$, as used in the base approach). We introduce an additional regression head for predicting the signed distance function, denoted as $g^i$. For labeled data pair as $(X_{l},Y) \in \mathcal{D}_l$, the supervised loss for segmentation task and regression task are: 
 \begin{align}
\mathcal{L}_{seg}^{l} &=  \ell_{ce}({P}_{l}^{1}, {Y}) + \ell_{ce}({P}_{l}^{2}, {Y}), \label{labeled superviseA}\\
\mathcal{L}_{sdf}^{l} &=  ||\phi_{pre}^{1} - \phi_{Y}  ||^{2} +||\phi_{pre}^{2} - \phi_{Y}  ||^{2}\label{labeled superviseB}
\end{align}
where $\ell_{ce}$ is the cross-entropy loss, $P^i_{l} = f^i(h^i(X_{l}))$ is the output of the segmentation head, $\phi^i_{pre} = g^i(h^i(X_{l}))$ is the output of the level set prediction head, and superscript $i\in[1,2]$ means different network. In other words, we complement the basic supervised loss given in Eq.~\eqref{labeled superviseA}, with a specific loss $\mathcal{L}_{sdf}^{l}$, given in \eqref{labeled superviseB}, dedicated to the signed distance function.

For unlabeled data $X_{u} \in D_{u}$, we denote the mutual supervision loss as:
 \begin{align}
\mathcal{L}_{seg}^{u} &=  \ell_{ce}({P}_{u}^{1}, \hat{Y}^2) + \ell_{dc}({P}_{u}^{2}, \hat{Y}^1), \label{unlabeled superviseA}\\
\mathcal{L}_{sdf}^{u} &=  ||\phi_{pre}^{1} - \phi_{pre}^{2} ||^{2}\label{unlabeled superviseB}
\end{align}
where $\hat{Y}^1$ and $\hat{Y}^2$ are pseudo labels obtained from $P_{u}^{1}$ and $P_{u}^{2}$, respectively. Again, we complement the base loss for unlabeled data, given in Eq.~\eqref{unlabeled superviseA}, with a specific cost, given in Eq.~\eqref{unlabeled superviseB} dedicated to the signed distance function. 

\begin{table*}[tb]
\caption{Quantitative comparison of  Dice score ($\uparrow$) between our proposed STAC and some class imbalance SSL segmentation methods on  \textbf{2\% labeled AMOS dataset}. 
}
\label{amos2}
\resizebox*{\linewidth}{!}{

\begin{tabular}{c|c|ccccccccccccccc}
\toprule
Methods  & Avg. Dice  & Sp   & RK   & LK   & {Ga}   & {Es}   & Li   & St   & Ao   & IVC   & PA   & {RAG}  & {LAG}  & Du   & Bl  & P/U  \\\midrule
 V-Net (fully)   &76.5   &92.2 &92.2  &93.3  &65.5  &70.3  &95.3  &82.4  &91.4  &85.0  &74.9 &58.6  &58.1  &65.6  &64.4  &58.3   \\ \midrule

 CReST~\cite{wei2021crest}      &32.1  &44.6  &45.6  &49.4  &18.3  &18.3  &52.5  &35.9   &35.9   &42.5   &25.4   &19.1   &10.5    &24.4   &37.0  &22.2  \\
 \rowcolor{gray!40}
  CReST+STAC      & 37.1\tiny{\textcolor{blue}{(+5.0)}}	&63.5   &58.5   &55.5  &32.6  &22.1   &69.2 &24.3 &49.1 &48.1 &18.6 &22.6 &13.1 &21.8 &32.9 &25.6  \\
\midrule

 SimiS~\cite{simis}     & 37.5	&62.4 &60.4 &59.5 &29.7 &0.0 &70.2 &37.1 &50.0 &46.2 &27.8 &21.5 &8.1 &21.0 &43.1 &26.3  \\
 \rowcolor{gray!40}
 SimiS+STAC     &38.0\tiny{\textcolor{blue}{(+0.5)}}	&59.6 &59.9 &58.9 &29.5 &0.0 &67.9 &39.6 &52.7 &50.6 &29.6 &19.4 &4.9 &26.5 &45.9 &25.6  \\
\midrule

 CLD~\cite{lin2022cld}        & 36.6	& 60.9 &54.1 &56.4 &28.1 &0.0 &67.7 &35.6 &48.1 &48.7 &29.7 &28.0 &6.8 &22.1 &40.5 &22.4   \\
\rowcolor{gray!40}
CLD+STAC        &40.0\tiny{\textcolor{blue}{(+3.4)}}	& 63.2 &61.4 &59.4 &32.4 &0.0 &73.2 &39.7 &59.3 &56.2 &37.0 &19.1 &8.1 &19.4 &46.4 &25.1   \\
\midrule

 DHC~\cite{wang2023dhc}   & 37.6   & 62.3	& 60.1	& 59.7	& 27.7	& 18.3	& 69.2	& 31.6	& 50.9	& 42.7	& 24.4	& 23.9	&9.5	& 13.5	& 46.7	& 23.7   \\
\rowcolor{gray!40}

DHC+STAC  &45.8\tiny{\textcolor{blue}{(+8.2)}}   &63.2 &63.6   &62.8   &38.4  &27.3 &73.5   &45.2 &66.5  &57.3  &41.5  &29.9  &13.7 &31.8   &45.6   &27.1  \\ \bottomrule
\end{tabular}

}

\end{table*}
\begin{table*}[tb]
\scriptsize
\caption{Quantitative comparison of  Dice score ($\uparrow$) between our proposed STAC and some class imbalance SSL segmentation methods on \textbf{5\% labeled AMOS dataset}. 
}
\label{amos5}
\resizebox*{1\linewidth}{!}{

\begin{tabular}{c|c|ccccccccccccccc}
\toprule
Methods  & Avg. Dice  & Sp   & RK   & LK   & {Ga}   & {Es}   & Li   & St   & Ao   & IVC   & PA   & {RAG}  & {LAG}  & Du   & Bl  & P/U  \\\midrule
 V-Net (fully)  &76.5   &92.2 &92.2  &93.3  &65.5  &70.3  &95.3  &82.4  &91.4  &85.0  &74.9 &58.6  &58.1  &65.6  &64.4  &58.3    \\ \midrule

 CReST~\cite{wei2021crest}       &47.0 &74.1 &66.7 &67.6 &24.1 &32.3 &86.1 &47.6 &72.4 &54.6 &39.5 &25.8 &13.4 &29.2 &44.6 &27.3 \\
 \rowcolor{gray!40}
  CReST+STAC     &48.0\tiny{\textcolor{blue}{(+1.0)}}  &73.6 &71.1 &70.2 &31.2 &32.3 &86.0 &44.0 &66.9 &57.5 &40.0 &24.2 &18.3 &29.7 &46.9 &28.2   \\
\midrule

 SimiS~\cite{simis}     &49.6   &80.1 &72.7 &69.5 &36.8 &34.9 &88.1 &53.4 &78.5 &58.3 &43.0 &29.3 &12.4 &21.9 &45.9 &19.6  \\
 \rowcolor{gray!40}
 SimiS+STAC     &53.7\tiny{\textcolor{blue}{(+4.1)}} &79.0 &77.4 &74.0 &41.0 &44.7 &88.3 &55.2 &78.8 &59.2 &37.5 &33.5 &28.3 &32.2 &48.0 &29.5 \\
\midrule

 CLD~\cite{lin2022cld}        &49.3	& 80.2 & 77.6 & 71.4 & 34.8 & 36.5 & 88.5 & 44.0 & 78.7 & 57.9 & 44.3 & 30.1 & 17.0 & 23.6 & 43.4 & 12.9   \\
\rowcolor{gray!40}
CLD+STAC        &50.9\tiny{\textcolor{blue}{(+1.6)}}	& 81.1 & 75.6 & 75.9 & 39.9 & 40.4 & 88.2 & 51.4 & 78.8 & 61.5 & 47.8 & 28.9 & 17.5 & 31.8 & 45.9 & 0.2   \\
\midrule

 DHC~\cite{wang2023dhc}   & 49.2 	& 80.6	& 68.3	&70.6	& 34.1	& 35.1	& 85.3	& 51.9	&73.4	& 59.0	&48.0	& 26.3	&16.0	&27.3	& 42.9	&20.7   \\
\rowcolor{gray!40}

DHC+STAC  & 53.9\tiny{\textcolor{blue}{(+4.7)}} & 79.9 & 78.0 & 76.3 & 38.5 & 45.9 & 85.8 & 53.6 & 76.4 & 62.0 & 47.3 & 32.1 & 27.1 & 38.1 & 47.8 & 20.0  \\ \bottomrule
\end{tabular}

}

\end{table*}
\begin{table}[tb]
\scriptsize
\caption{Quantitative comparison of  average surface distance (ASD $\downarrow$) between our proposed STAC and some class imbalance SSL segmentation methods.
}
\label{amos2_asd}
\centering
\resizebox*{0.7\linewidth}{!}{

\begin{tabular}{c|cc|cc}
\toprule
\multirow{2}{*}{Methods}  & \multicolumn{2}{c|}{Amos}  & \multicolumn{2}{c}{Synapse}   \\ 
\cline{2-3}  \cline{4-5}
 &\raisebox{-1.5pt}{2\%}   &\raisebox{-1.5pt}{5\%}     &\raisebox{-1.5pt}{10\%}   &\raisebox{-1.5pt}{20\%}\\  \midrule

 CReST~\cite{wei2021crest}      &22.5	&9.7 	&50.9  	&29.6   \\
 \rowcolor{gray!40}
  CReST + STAC      &19.3	&5.9    &31.0  	&14.3  \\
\midrule

 SimiS~\cite{simis}     &26.3	&10.9   &32.9  &38.4\\
 \rowcolor{gray!40}
 SimiS + STAC     &26.2	 &5.6  &31.9   &32.7\\
\midrule

 CLD~\cite{lin2022cld}     &28.6  &11.4  &37.9  &20.2\\
\rowcolor{gray!40}
CLD + STAC        &26.0  &13.6   &22.8    &17.1\\
\midrule

 DHC~\cite{wang2023dhc}  &20.1 &11.2  &25.8  &9.8 \\
\rowcolor{gray!40}

DHC + STAC   &13.5  &6.4   &26.7  &8.9
\\ \bottomrule
\end{tabular}

}

\end{table}

\begin{table}[tb]
  \centering
  \caption{Quantitative results of the proposed STAC compared with other shape transformation on AMOS dataset using 2\% labeled images under the DHC~\cite{wang2023dhc} baseline. }
  \label{tab:other_transform}
  \setlength{\tabcolsep}{1mm}
   \resizebox{1\linewidth}{!}{
  \begin{tabular}{l  @{\hspace{1cm}}c  @{\hspace{1cm}}c  @{\hspace{1cm}}c @{\hspace{1cm}}c }

    \toprule
           &DHC~\cite{wang2023dhc} &+Elastic &+Anatomy~\cite{kovacs2023anatomy} & +STAC   \\
    \midrule
        Dice (\%)  &37.6   & 40.1 &40.0  & 45.8  \\
        ASD   &20.1  & 20.7 &16.6  & 13.5  \\
    \bottomrule

   \end{tabular} 
    }
\end{table}

\begin{table}[tb]
  \centering
  \caption{Quantitative results of applying STAC on labeled images (L) and unlabeled images (U) on AMOS dataset using 2\% labeled images under the DHC~\cite{wang2023dhc} baseline.}
  \label{tab:labeled_transform}
  \setlength{\tabcolsep}{1mm}
   \resizebox{0.9\linewidth}{!}{
  \begin{tabular}{l  @{\hspace{1cm}}c  @{\hspace{1cm}}c  @{\hspace{1cm}}c  }

    \toprule
      &DHC~\cite{wang2023dhc} & +STAC (L) &+STAC (U+L) \\
    \midrule
    Dice (\%)  &37.6  &41.7   & 45.8   \\
    ASD   &20.1  &17.1    &13.5  \\
    \bottomrule

   \end{tabular} 
    }
\end{table}
\begin{table*}[tb]
\scriptsize
\caption{Quantitative comparison of  Dice score ($\uparrow$) between our proposed STAC and some class imbalance SSL segmentation methods on \textbf{10\% labeled Synapse dataset}. 
}
\label{synapse10}
\resizebox*{\linewidth}{!}{
\begin{tabular}{c|c|ccccccccccccc}
\toprule
Methods  & Avg. Dice     & Sp   & RK   & LK   & {Ga}   & {Es}   & Li   & St   & Ao   & IVC  & PSV  & PA   & RAG  & LAG  \\\midrule
 V-Net (fully)    &62.0   &84.6 &77.2  &73.8  &73.3  &38.2  &94.6  &68.4  &72.1   &71.2  &58.2  &48.5  &17.9  &29.0 \\ \midrule

CReST~\cite{wei2021crest}   &24.3                  &39.6 &44.7 &26.3 &26.3 &1.4 &59.3 &16.2 &38.5 &30.2 &3.7 &8.1 &20.1 &1.9  \\
\rowcolor{gray!40}
CReST+STAC   &26.9\tiny{\textcolor{blue}{(+2.6)}}  &46.9 &49.3 &36.5 &30.2 &9.6 &51.5 &27.4 &42.5 &23.0 &2.9 &3.7 &23.5 &3.8   \\
\midrule

SimiS~\cite{simis}   &24.5                             &43.1 &38.0 &43.8 &16.5 &4.5 &68.1 &21.8 &46.3 &14.4 &4.7 &7.6 &0.5 &9.5   \\
\rowcolor{gray!40}
SimiS+STAC   &29.1\tiny{\textcolor{blue}{(+4.6)}} &48.1 &43.8 &42.4 &18.3 &9.1 &54.1 &23.7 &59.6 &50.9 &1.6 &11.6 &2.0 &14.0  \\
\midrule
                                 
CLD~\cite{lin2022cld}  &24.1    &48.1 &45.9 &39.1 &1.0 &9.4 &46.8 &19.4 &52.6 &33.3 &8.2 &2.4 &5.3 &1.9   \\
\rowcolor{gray!40}
CLD+STAC    &29.7\tiny{\textcolor{blue}{(+5.6)}}  &53.1 &55.0 &45.8 &0.4 &9.7 &54.1 &35.7 &50.0 &49.2 &8.1 &10.4 &0.6 &14.6  \\
\midrule
                                  
DHC ~\cite{wang2023dhc} &30.0		&60.7 &43.9 &42.4 &0.0 &19.2 &61.2 &36.9 &48.8 &38.3 &9.0 &11.8 &13.2 &4.9  \\ 
\rowcolor{gray!40}
DHC+STAC   &34.8\tiny{\textcolor{blue}{(+4.8)}}  &61.4 &61.5 &60.3 &0.2 &13.9 &74.3 &35.1 &61.0 &53.1 &11.6 &10.7 &9.5 &0.0\\ 
\bottomrule
\end{tabular}
}
\end{table*}
\begin{table*}[tb]
\scriptsize
\caption{Quantitative comparison of  Dice score ($\uparrow$) between our proposed STAC and some class imbalance SSL segmentation methods on \textbf{20\% labeled Synapse dataset}.
}
\label{synapse20}
\resizebox*{\linewidth}{!}{
\begin{tabular}{c|c|ccccccccccccc}
\toprule
Methods  & Avg. Dice     & Sp   & RK   & LK   & {Ga}   & {Es}   & Li   & St   & Ao   & IVC  & PSV  & PA   & {RAG}  & {LAG}  \\\midrule
 V-Net (fully)  &62.0   &84.6 &77.2  &73.8  &73.3  &38.2  &94.6  &68.4  &72.1  
 &71.2  &58.2  &48.5  &17.9  &29.0    \\ \midrule

CReST~\cite{wei2021crest}   &36.5                      &68.5 &66.7 &53.1 &36.9 &3.0 &83.5 &25.7 &40.6 &39.8 &9.0 &15.8 &14.3 &17.6    \\
\rowcolor{gray!40}
CReST+STAC   &38.4 \tiny{\textcolor{blue}{(+1.9)}}  &62.3 &70.5 &58.8 &27.0 &19.4 &76.0 &33.8 &52.7 &37.8 &2.3 &15.5 &17.2 &26.1   \\
\midrule

SimiS~\cite{simis}   &41.8  &84.6 &78.5 &74.7 &2.9 &0.0 &82.5 &44.0 &66.3 &64.2 &29.1 &17.7 &0.0 &0.0    \\
\rowcolor{gray!40}
SimiS+STAC    &43.5 \tiny{\textcolor{blue}{(+1.7)}} &83.6 &77.4 &76.2 &4.3 &0.0 &81.2 &43.2 &74.5 &68.4 &30.0 &23.6 &3.7 &0.0 \\
\midrule
                                 
CLD~\cite{lin2022cld}  &45.0    &76.6 &76.9 &72.3 &9.9 &0.0 &87.4 &34.1 &63.3 &64.7 &14.5 &19.4 &29.1 &37.7    \\
\rowcolor{gray!40}
CLD+STAC    &46.2 \tiny{\textcolor{blue}{(+1.2)}}   &82.9 &79.4 &79.4 &17.4 &0.0 &87.5 &19.1 &72.3 &61.4 &20.0 &24.4 &27.9 &29.7  \\
\midrule
                                  
DHC ~\cite{wang2023dhc}  &47.8  &75.5 &73.3 &77.0 &1.7 &24.5 &80.8 &38.8 &65.6 &57.2 &28.4 &20.6 &25.7 &52.2 \\ 
\rowcolor{gray!40}
DHC+STAC  &50.7 \tiny{\textcolor{blue}{(+2.9)}} &84.2 &81.4 &80.4 &6.5 &26.3 &84.8 &33.2 &76.4 &65.8 &38.3 &21.9 &22.8 &38.0  \\ 

\bottomrule
\end{tabular}
}
\end{table*}

In summary, our proposal boils down to adding to a base network, another head to predict the signed distance function, together with a specific cost function for training this head.
During the training process, we use $X^{transformed}$ (Eq.~\eqref{shape_transform_image}) and $Y^{transformed}$ (Eq.~\eqref{shape_transform_label}) as  data-augmentation (see Fig.~\ref{fig:pipeline}).

\begin{figure*}[tb]
\centering
\subfigure[GT]{
\begin{minipage}[b]{0.13\linewidth}
\includegraphics[width=1\linewidth]{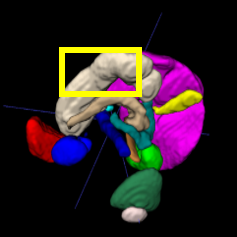}\vspace{1pt}
\includegraphics[width=1\linewidth]{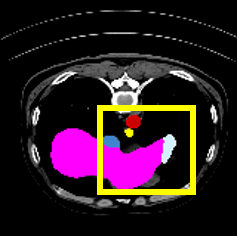}\vspace{1pt}
\includegraphics[width=1\linewidth]{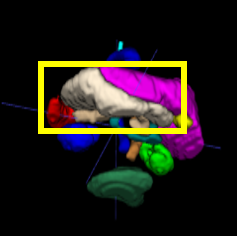}\vspace{1pt}
\includegraphics[width=1\linewidth]{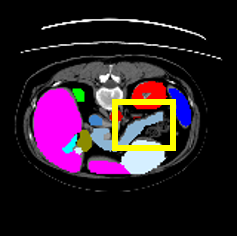}
\end{minipage}}\hspace{-1.5mm}
\subfigure[DHC+STAC]{
\begin{minipage}[b]{0.13\linewidth}
\includegraphics[width=1\linewidth]{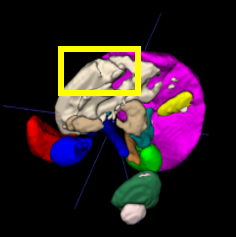}\vspace{1pt}
\includegraphics[width=1\linewidth]{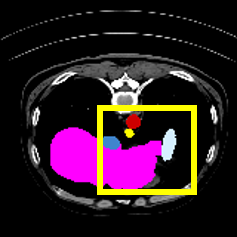}\vspace{1pt}
\includegraphics[width=1\linewidth]{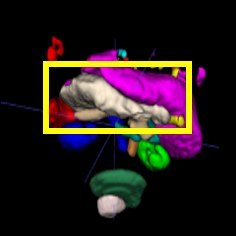}\vspace{1pt}
\includegraphics[width=1\linewidth]{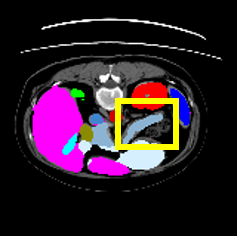}
\end{minipage}}\hspace{-1.5mm}
\subfigure[DHC]{
\begin{minipage}[b]{0.13\linewidth}
\includegraphics[width=1\linewidth]{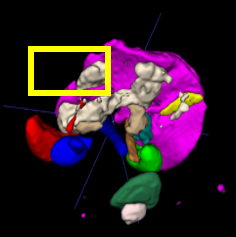}\vspace{1pt}
\includegraphics[width=1\linewidth]{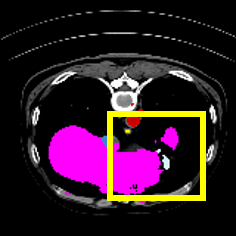}\vspace{1pt}
\includegraphics[width=1\linewidth]{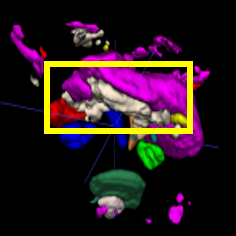}\vspace{1pt}
\includegraphics[width=1\linewidth]{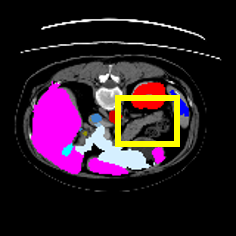}
\end{minipage}}\hspace{-1.5mm}
\subfigure[CLD]{
\begin{minipage}[b]{0.13\linewidth}
\includegraphics[width=1\linewidth]{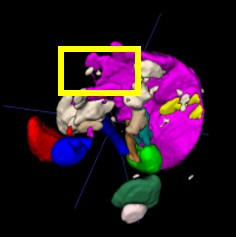}\vspace{1pt}
\includegraphics[width=1\linewidth]{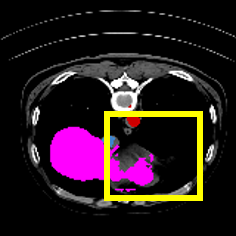}\vspace{1pt}
\includegraphics[width=1\linewidth]{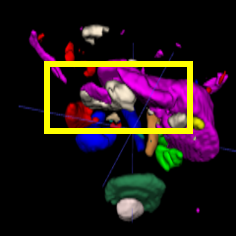}\vspace{1pt}
\includegraphics[width=1\linewidth]{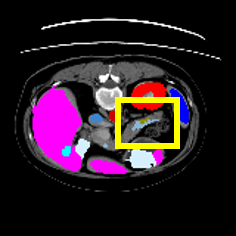}
\end{minipage}}\hspace{-1.5mm}
\subfigure[CReST]{
\begin{minipage}[b]{0.13\linewidth}
\includegraphics[width=1\linewidth]{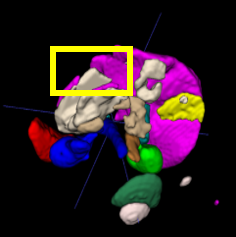}\vspace{1pt}
\includegraphics[width=1\linewidth]{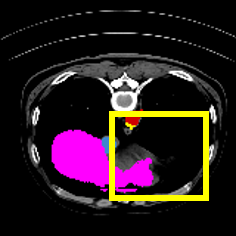}\vspace{1pt}
\includegraphics[width=1\linewidth]{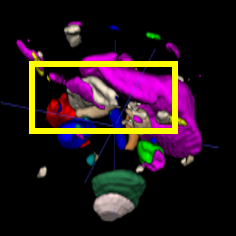}\vspace{1pt}
\includegraphics[width=1\linewidth]{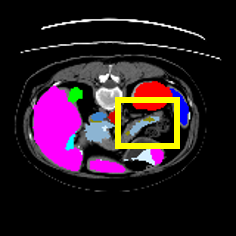}
\end{minipage}}\hspace{-1.5mm}
\subfigure[SimiS]{
\begin{minipage}[b]{0.13\linewidth}
\includegraphics[width=1\linewidth]{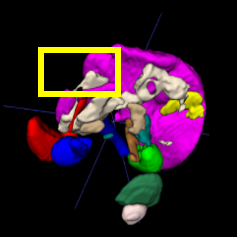}\vspace{1pt}
\includegraphics[width=1\linewidth]{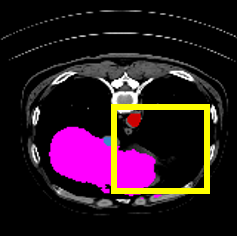}\vspace{1pt}
\includegraphics[width=1\linewidth]{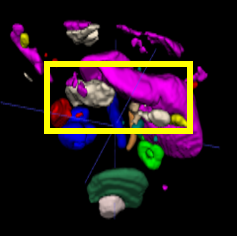}\vspace{1pt}
\includegraphics[width=1\linewidth]{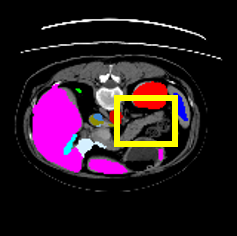}
\end{minipage}}\hspace{-1.5mm}

\caption{Some qualitative segmentation results of STAC and some other methods on AMOS dataset. The first and third rows are 3D views, and the second and fourth rows are 2D slices. }
\label{fig:view}
\end{figure*}

\section{Experiments}
\subsection{Dataset and Implementation Details.}

To address class imbalance in semi-supervised medical image segmentation, we adhere to the experimental setup proposed in DHC~\cite{wang2023dhc}. We use two different benchmarks: the Synapse\footnote{\url{https://www.synapse.org/\#!Synapse:syn3193805/wiki/89480}}~\cite{synapse}  and AMOS\footnote{\url{https://amos22.grand-challenge.org/}}~\cite{amos} datasets. The AMOS dataset consists of 360 scans, which are divided into 216, 24 and 120 scans for training, validation, and testing, covering 15 foreground classes, including  spleen (Sp), right kidney (RK), left kidney (LK), gallbladder (Ga), esophagus (Es), liver (Li), stomach (St), aorta (Ao), inferior vena cava (IVC), pancreas (PA), right adrenal gland (RAG), left adrenal gland (LAG), duodenum (Du), bladder (Bl), prostate/uterus (P/U).
Compared with AMOS, Synapse excludes Du, Bl and P/U but adds portal \& splenic veins (PSV). 
Synapse dataset consists of 30 scans, and we follow DHC~\cite{wang2023dhc} splitting them as 20, 4 and 6 scans for training, validation, and testing, respectively.  The proposed method is evaluated with two widely used metrics in semi-supervised medical image segmentation: Dice coefficient (Dice) and the average surface distance (ASD).

\subsection{Comparison with State-of-the-Art Methods.}
To validate the effectiveness of our proposed Shape Transformation driven by Active Contour (STAC) method, we conduct experimental evaluations by integrating STAC with four established baselines: CReST~\cite{wei2021crest}, SimiS~\cite{simis}, CLD~\cite{lin2022cld}, and DHC~\cite{wang2023dhc}, which address the class imbalance problem. Following the experiments setting in DHC~\cite{wang2023dhc}, we conduct semi-supervised experiments on the AMOS dataset using 2\% and 5\% of the labeled data, and experiments on the Synapse dataset using 10\% and 20\% of the labeled data. The results of the AMOS dataset~\cite{amos} experiments are depicted in Tab.~\ref{amos2} (2\%) and Tab.~\ref{amos5} (5\%), and the results of the Synapse dataset~\cite{synapse} experiments are presented in Tab.~\ref{synapse10} (10\%) and Tab.~\ref{synapse20} (20\%). Under the setting of using 2\% labeled data of AMOS dataset~\cite{amos}, applying our STAC to the state-of-the-art DHC~\cite{wang2023dhc}  method resulted in a 8.2\% improvement in the Dice coefficient. In the scenario where 5\% of the AMOS dataset is labeled, the implementation of our STAC on DHC~\cite{wang2023dhc}  led to an enhancement of 4.7\% in the Dice coefficient. On the Synapse dataset~\cite{synapse}, when using 10\% labeled data, applying our method to the four approaches—CReST~\cite{wei2021crest} , SimiS~\cite{simis}, CLD~\cite{lin2022cld}, and DHC~\cite{wang2023dhc}—results in Dice coefficient improvements of 2.6\%, 4.6\%, 5.6\%, and 4.8\%, respectively.
Some qualitative results are shown in Fig.~\ref{fig:view}.

\subsection{Ablation Studies.}
\noindent
\textbf{Ablation study using other shape transformation.} While random elastic~\cite{connor2019survey} is a simple shape transformation augmentation, it cannot be used to enlarge organs. Tab.~\ref{tab:other_transform} shows that, while random elastic marginally enhances the Dice coefficient, it increases ASD by altering organ contours. STAC, by evolving along the normal vector to avoid distorted shapes, produces more realistic and stable shape transformations. This leads to a 5.8\% improvement in Dice scores over the anatomy-informed shape transformation method~\cite{kovacs2023anatomy}.

\begin{figure}[tb]
\centering
\includegraphics[width = 0.7\linewidth]{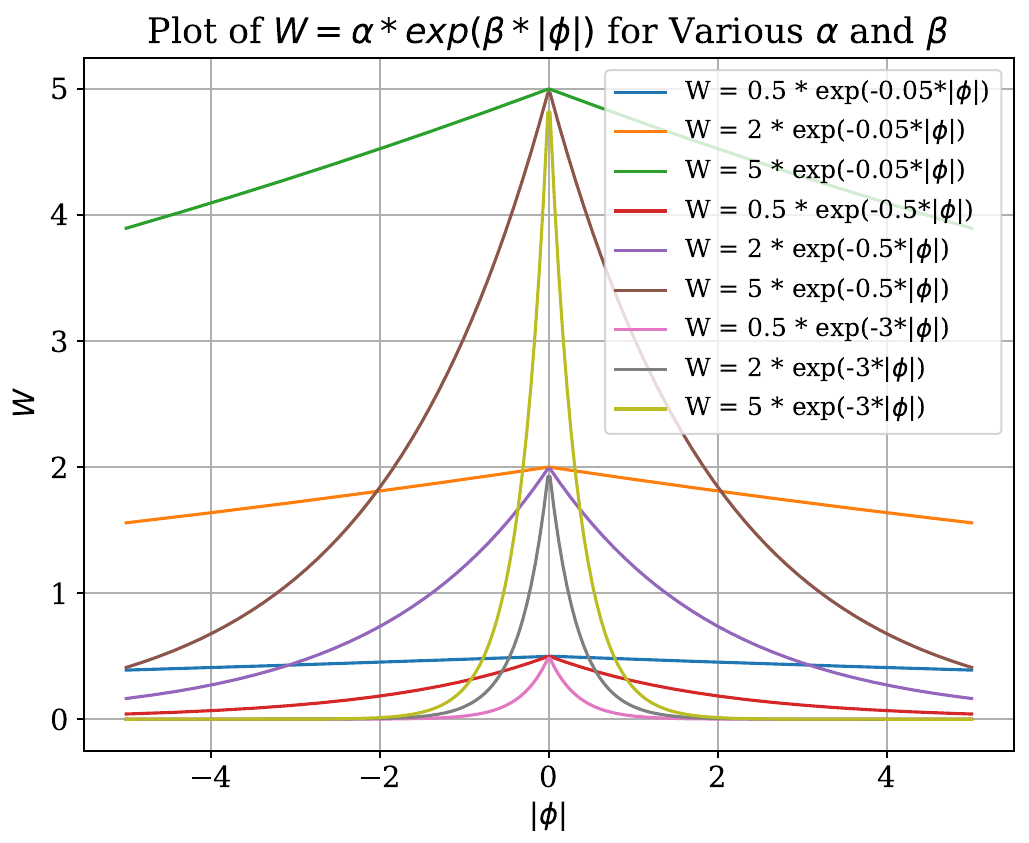}
\caption{Plot of the SDF-based weight function with different parameters $\alpha$ and $\beta$.
} 
\label{fig:plot}
\end{figure}

\begin{figure}[tb]
\centering
\includegraphics[width = 0.7\linewidth]{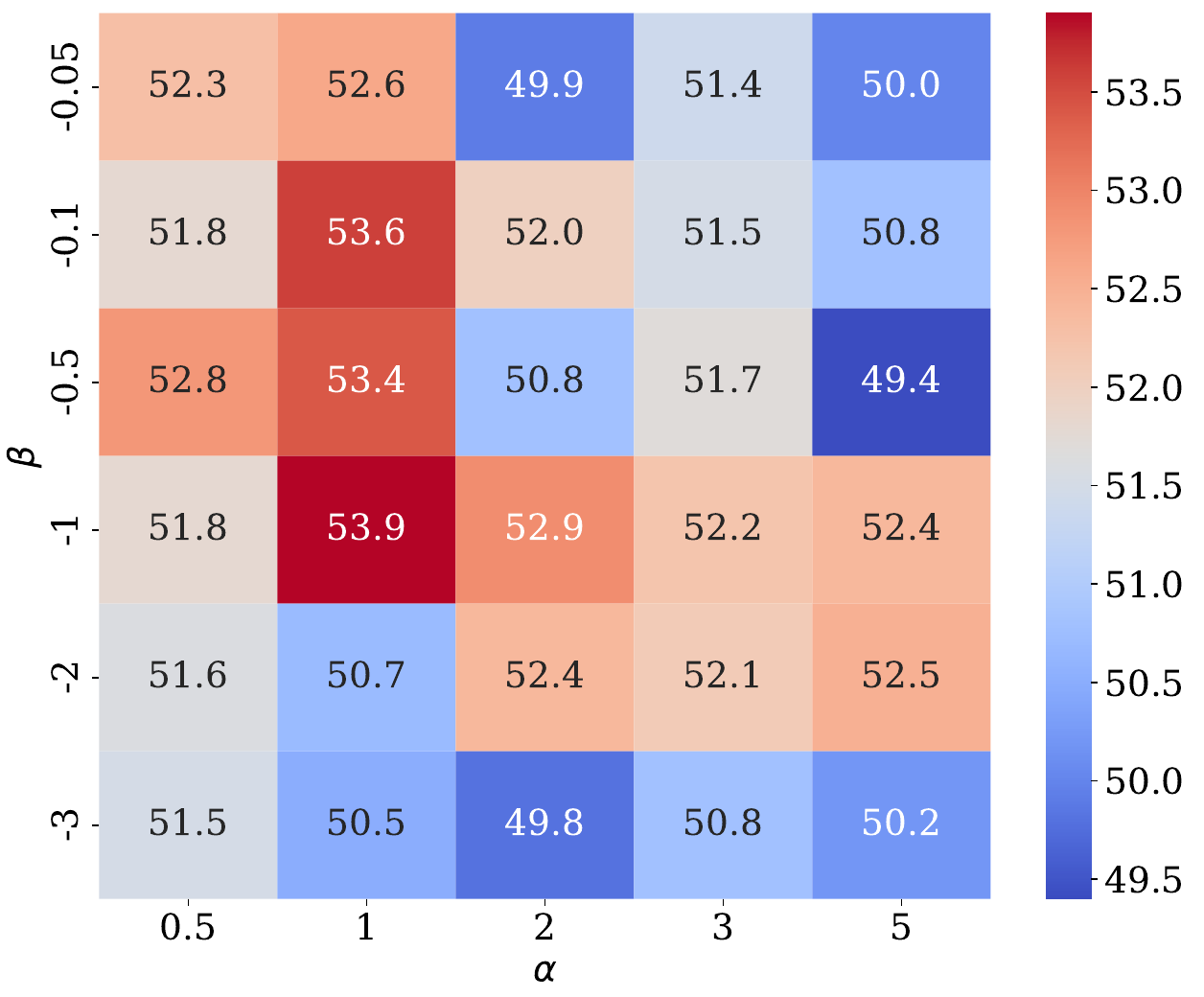}
\caption{Dice score of using different values for parameters $\alpha$ and $\beta$ on AMOS dataset using 5\% labeled images under the DHC~\cite{wang2023dhc} baseline. The numerical values in the heatmap represent Dice scores.
} 
\label{fig:ablation}
\end{figure}

\noindent  \textbf{Ablation study on hyperparameters ($\alpha$ and $\beta$) of $W_{\mathcal{O}}$.} The proposed STAC only include $\alpha$ and $\beta$ as hyperparameters. The curve of the SDF-based weight function with different parameters $\alpha$ and $\beta$ is shown in Fig.~\ref{fig:plot}. The experimental results obtained by controlling the degree of shape change using the SDF-based weight function are shown in Fig.~\ref{fig:ablation}. The best results are achieved when $\alpha = 1$ and $\beta = -1$, with a Dice score of 53.9.

\noindent  \textbf{Ablation study focusing on the application of shape transformation exclusively to labeled data.}  
Our method can be applied solely to the images and annotation pairs of labeled samples, which does not alter the training process. As shown in Tab.~\ref{tab:labeled_transform}, performing shape transformations only on labeled samples also significantly improves the results.

\section{Conclusion}
This paper introduces a novel Shape Transformation driven by Active Contour (STAC) method for addressing the issue of imbalanced class distribution in 3D medical image segmentation. By utilizing the Signed Distance Function (SDF) as the level set function, and 
deforming the voxels in the direction of the steepest descent of SDF, STAC effectively enlarges smaller organs, balancing the class distribution across various organs. This approach ensures that voxels distant from the expanding organs remain unchanged by designing an SDF-based weight function to control the degree of deformation.  Experimental results show that using STAC as a data-augmentation process in four baseline models yields consistent and significant improvements, demonstrating the effectiveness of our method.

\section{Acknowledgement}
This work was supported in part by the National Key Research and Development Program of China (2023YFC2705700), NSFC 62222112 and 62176186,  the NSF of Hubei Province of China (2024AFB245).

\bibliographystyle{IEEEtran}
\bibliography{ref.bib}

\end{document}